\documentclass[conference]{IEEEtran}
\IEEEoverridecommandlockouts
\usepackage{cite}
\usepackage{amsmath,amssymb,amsfonts}
\usepackage{algorithmic}
\usepackage{graphicx}
\usepackage{textcomp}
\usepackage{xcolor}
\usepackage{svg}
\usepackage{enumitem}

\usepackage[utf8]{inputenc}
\usepackage[ruled,vlined,linesnumbered]{algorithm2e}
\usepackage{xcolor}
\usepackage{rotating,color}
\usepackage{makecell}

\usepackage{caption}
\usepackage{subcaption}

\SetCommentSty{mycommfont}
\SetKwInput{KwInput}{Input} 
\SetKwInput{KwOutput}{Output} 

\def\BibTeX{{\rm B\kern-.05em{\sc i\kern-.025em b}\kern-.08em
    T\kern-.1667em\lower.7ex\hbox{E}\kern-.125emX}}

\begin{document}

\title{A Surrogate Model Framework for Explainable Autonomous Behaviour}

\author{\IEEEauthorblockN{Konstantinos Gavriilidis}
\IEEEauthorblockA{\textit{Heriot-Watt University} \\
Edinburgh, United Kingdom \\
kg47@hw.ac.uk}
\and
\IEEEauthorblockN{Andrea Munafo}
\IEEEauthorblockA{\textit{SeeByte Ltd.} \\
Edinburgh, United Kingdom \\
andrea.munafo@seebyte.com}
\and
\IEEEauthorblockN{Wei Pang}
\IEEEauthorblockA{\textit{Heriot-Watt University} \\
Edinburgh, United Kingdom \\
W.Pang@hw.ac.uk}
\and
\IEEEauthorblockN{Helen Hastie}
\IEEEauthorblockA{\textit{Heriot-Watt University} \\
Edinburgh, United Kingdom \\
H.Hastie@hw.ac.uk}
}

\maketitle

\begin{abstract}
Adoption and deployment of robotic and autonomous systems in industry are currently hindered by the lack of transparency, required for safety and accountability. Methods for providing explanations are needed that are agnostic to the underlying autonomous system and easily updated. Furthermore, different stakeholders with varying levels of expertise, will require different levels of information. In this work, we use surrogate models to provide transparency as to the underlying policies for behaviour activation.  We show that these surrogate models can effectively break down autonomous agents' behaviour into explainable components for use in natural language explanations. 
\end{abstract}



\section{Introduction}

Robotic and autonomous systems are at the stage where we are seeing them being adopted more frequently in a variety of environments, such as ground vehicles for inspection of disaster sites, or underwater for pipeline inspection. It is important that humans are kept in the loop in these operations and are able to intervene as necessary. However, this comes with challenges, such as underwater vehicles having limited bandwidth to broadcast updates \cite{b1}. Transparency and the ability to explain actions and decisions are key factors for safety, accountability and adoption \cite{b4}. However, these are non-trivial to implement, given the complexity of autonomous systems and the `blackbox' nature of neural-based models. 

Platform-specific explanation interfaces normally require a basic understanding of an agent's behaviour space (\begin{math} B\end{math}), possible states (\begin{math} S\end{math}) and decision-making (\begin{math} D\end{math}) to comprehend its capabilities and what could be described to operators. Furthermore, user studies are necessary to recognise which behaviours may be certainly valid and appropriate but might perhaps confuse the operator. This can lead to mission aborts and inaccurate mental models \cite{b3}. 

These user studies and explanation methods also need to adapt to the stakeholder. The IEEE Standard for Transparency of Autonomous Systems (P7001) \cite{b4}, defines a number of stakeholder groups from the expert operator, to the general public and lawmakers. They all require different types of information and level of detail to be included in the explanation. 
 
Furthermore when working with commercial entities, they are continuously developing their autonomous models, adding new behaviours or states as required for new use cases and customers. If we are to provide accurate, up-to-date explanations, the explanation module will also need continuous updating, requiring considerable time and effort.

Here, we propose a generic  method using surrogate models for generating autonomy-agnostic explanations that can be used without a deep understanding of the underlying autonomy and can be easily updated. Specifically addressing the following research questions:

\begin{itemize} 
\label{research_questions}
    \item \textbf{RQ1:} How robust are surrogate models in approximating a complex deterministic agent's policy for behaviour activation?
     \item \textbf{RQ2:} Can these surrogate models be used to effectively generate explanations? 
    \item \textbf{RQ3:} How is the performance affected when going from simulated data to real trials with real vehicles tested in a realistic environment?
\end{itemize}

\begin{figure}[t!]
\centering
\includegraphics[width=1.0\columnwidth]{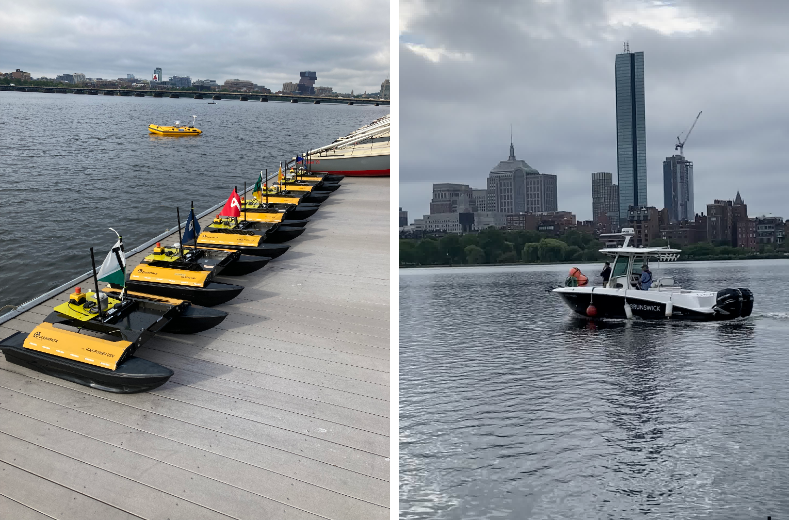}
\caption{Unmanned Surface Vehicles (USVs) Heron (left) and Philos (right) used during the trials on Charles River in Boston.}
\label{fig: unmanned_surface_vehicles}
\end{figure}

For the remainder of this paper, in Section \ref{related_work} we mention previous work, which has impacted our approach and in Section \ref{use_case_and_exp} we describe our use case and the explanation types that our framework provides. In Section \ref{system_overview}, we describe the functionality of components, such as the surrogate models or the feature contribution estimators in the pipeline architecture. Finally, in Section \ref{results}, we report the performance of the surrogate models in simulations and how it is affected during the trial or when new behaviours are incremented. 


\begin{figure}[t!]
\centering
\includegraphics[width=0.6\columnwidth]{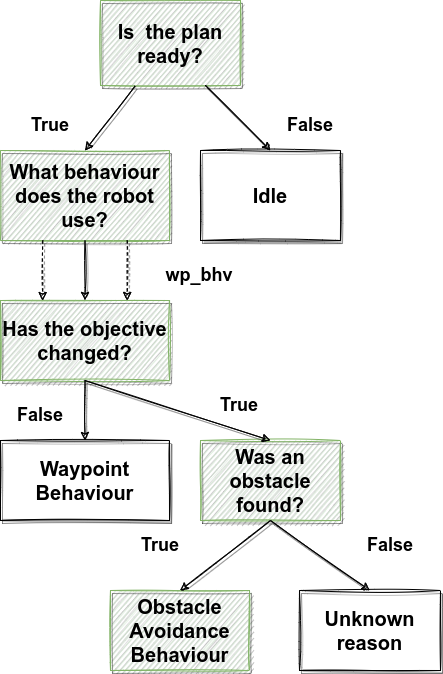}
\caption{Empirical decision tree for behaviour activation derived by a domain expert.}
\label{fig: empirical_decision_tree}
\end{figure}

\section{Related Work} \label{related_work}

\textbf{Explanation Frameworks:}  Explainable agency has been introduced as a trait of robots to define what properties a transparent robot should have conceptually. Among these traits are the ability to explain (i) plan generation, (ii) executed actions and (iii) replanning in a user-friendly fashion \cite{b5}.   Looking at previous work on explainable agency, the whole process can be broken down into three main parts, \textit{explanation generation}, \textit{explanation communication} and \textit{explanation reception} \cite{b6}. For explanation communication, previous studies have looked into how robots could explain themselves as people do by including reasons for intentional and causes for unintentional behaviours \cite{b7}. Further studies investigate the desired verbosity of explanations in different scenarios \cite{b8,b9} and various types of explanations, which can be provided to a user from an explainable planning perspective \cite{b10}. 

\textbf{Explainable Artificial Intelligence:} The right of users to receive explanations about a robot's behaviour is supported by government regulations and the recommended direction is the development of transparent robotics \cite{b4}, e.g. through the use of interpretable models \cite{b11}. Machine learning models differ in terms of simulatability, decomposability and algorithmic transparency \cite{b12}. In the case of opaque models, explanation methods should be applied to them to disambiguate their functionality. Explanation methods fall into two categories depending on the way they are applicable to a black-box model, namely \textit{Model-specific} and \textit{Model-agnostic} \cite{b13}. Surrogate models can belong to either category, depending on their intended use and are useful for deriving the causality behind any prediction. LIME \cite{b14} accomplishes this by locally approximating the model around a given prediction. SHAP \cite{b15} generates Shapley values, which indicate the contribution of each feature to the difference between the initial belief of a model and its actual prediction. Another option to highlight causal relationships is through counterfactuals, where several feature contributions can help the user understand the relationship between feature values and the corresponding predictions \cite{b16}. However, surrogate models are not always consistent and, as a result, robustness has been introduced as a metric that represents how stable explanations are when inputs are slightly modified \cite{b17}.

\textbf{Explainable Robotics:} In Sakai and Nagai \cite{b18}, the relationship between XAI and explanations for transparent robotics is defined. Existing work has examined the use of both algorithmically transparent models and the combination of opaque models with posthoc explanation methods to explain robotic failures \cite{b19}. The decision-making of reinforcement learning agents has also been explained either with the use of surrogate models \cite{b20, b21} or the generation of Shapley values to explain robot grasping \cite{b22}. Focusing more on explanation communication and reception, the studies in Thielstrom et al. \cite{b23} and Robb et al. \cite{b24} present videos of robotic failures along with explanations to users. Meanwhile, in Das et al. \cite{b25}, natural language explanations are generated with a Neural Translation Network to improve human assistance in fault recovery. Each approach performed a corresponding user study to evaluate how explanations affect the mental model of users.   

Continuing our effort in \cite{b26}, we have developed a framework that retrieves data from deterministic agents and, with model selection, finds the optimal classifier for behaviour prediction. Depending on the transparency of the intermediate model, we capture the causality behind behaviours either by directly analysing the model or with the application of a posthoc explainer. 

\textbf{Knowledge Representation and Verbalisation:} Knowledge representations have always played an important role in the unification or the completion of knowledge for autonomous agents. In Li et al. \cite{b27}, an ontology is used to tackle the issue of information heterogeneity  and to facilitate collaboration between underwater robots.
Additionally, in Gavriilidis et al. \cite{b28} an ontology is used to relate sensor readings to hardware errors and make a new plan, while a ROS listener retrieves and verbalises these outcomes with a surface realiser. Furthermore, Suh et al. \cite{b29} use a multilayered ontology to complement the perception of a household robot for  object recognition and assist with its localisation. At the same time, knowledge representations play an important role in Natural Language Generation. In \cite{b30}, a Neural Language Model efficiently transforms Wikipedia infoboxes into biography summaries, while in \cite{b31} a fine-tuned T5 model generates sentences just by connecting plain utterances from concept sets. On the other hand, Ghosal \cite{b32} collected a dialogue reasoning dataset, where additional context is incorporated into utterances to teach a T5 model to make more intuitive transitions in dialogue. This type of data-driven natural language generation is out of scope for the work described here, but is clearly an area for future use, particularly with the advent of more advanced large language models such as GPT-4 \cite{b33}. 
   

\section{Use Case and Explanation Types}
\label{use_case_and_exp}


The use case examined here is a hybrid autonomy, that combines a ROS-based deterministic agent with a reactive agent, prioritising behaviours through multi-objective optimization for the maritime domain. Specifically, our scenarios focus on \textit{Unmanned Surface Vehicles} (USV) and \textit{Autonomous Underwater Vehicles} (AUV), as illustrated in Figure \ref{fig: unmanned_surface_vehicles}. This work was done in collaboration with industry partner SeeByte Ltd, who have developed an autonomous agent for driving such vehicles for a variety of maritime applications, such as inspection. 
To have an initial understanding of the autonomous agent and how it selects a behaviour, we interviewed in-depth a domain expert from the company. From this  interview, we derived an abstract definition of behaviour decision-making in a tree format. This \textit{empirical decision tree} is illustrated in Figure \ref{fig: empirical_decision_tree}.

We then investigated if this behaviour tree covers aspects of a mission in simulation ahead of the real trial.  The simulation scenario involves a restricted coastal area on the River Charles in Boston, where two vehicles collaborate to complete a mission. The versatile USV Heron performs each objective according to the uploaded plan, while USV Philos inspects the area to detect obstacles and notifies the other vehicle if something out of the plan is found. Each mission contains a launch and recovery objective and a number of survey or target objectives, where the vehicle needs to hold its position for a default amount of time. Additionally, the obstacles are either static (with locations specific to each mission but there for the duration), or dynamic (i.e. appearing during the mission). In terms of capabilities, both vehicles support 6 behaviours  \textit{B = (wait, transit, survey, hold\_position, replanned\_transit, avoid\_obstacle)} and they both run two autonomy models simultaneously (one for each vehicle) in a master-slave architecture. 

\label{explanation_types}
The following explanation types were derived in consultation with the expert and captured in the empirical decision tree in Figure \ref{fig: empirical_decision_tree} and are listed here.

\begin{enumerate}[label=\textbf{E.\arabic*}]
    \item \textbf{Behaviour Causality} describes how a robot selects its current behaviour or action. Especially for operators, it can be difficult to comprehend how a robot closely observes objects around it, updates its world model and acts according to its goals. The utterances of this explanation usually entail the name of the behaviour, its use and the cause of activation. \textit{\textbf{Answers question:} Why did you do that?} \label{1}
    \item \textbf{Replanning Clarification} complements the previous category and covers cases where unexpected outcomes arise and force the autonomy to alter its plan. Some indicative examples are obstacle avoidance and platform integrity, where for safety reasons the robot has to make a stop in a new location. \textit{\textbf{Answers question:} Why do I need to replan at this point?} \label{2}
    \item \textbf{Counterfactual Explanation} allows the operator to ask the autonomy how it would react if its internal state changed in a specific way. With this functionality, the user can learn about alternative outcomes at any given point and to better comprehend the underlying logic of the autonomous agent. \textit{\textbf{Answers question:} What if?}\label{3} 
\end{enumerate}

\section{Method}
\label{system_overview}

\begin{figure*}[!ht]
\centering
\includegraphics[width=1.0\textwidth]{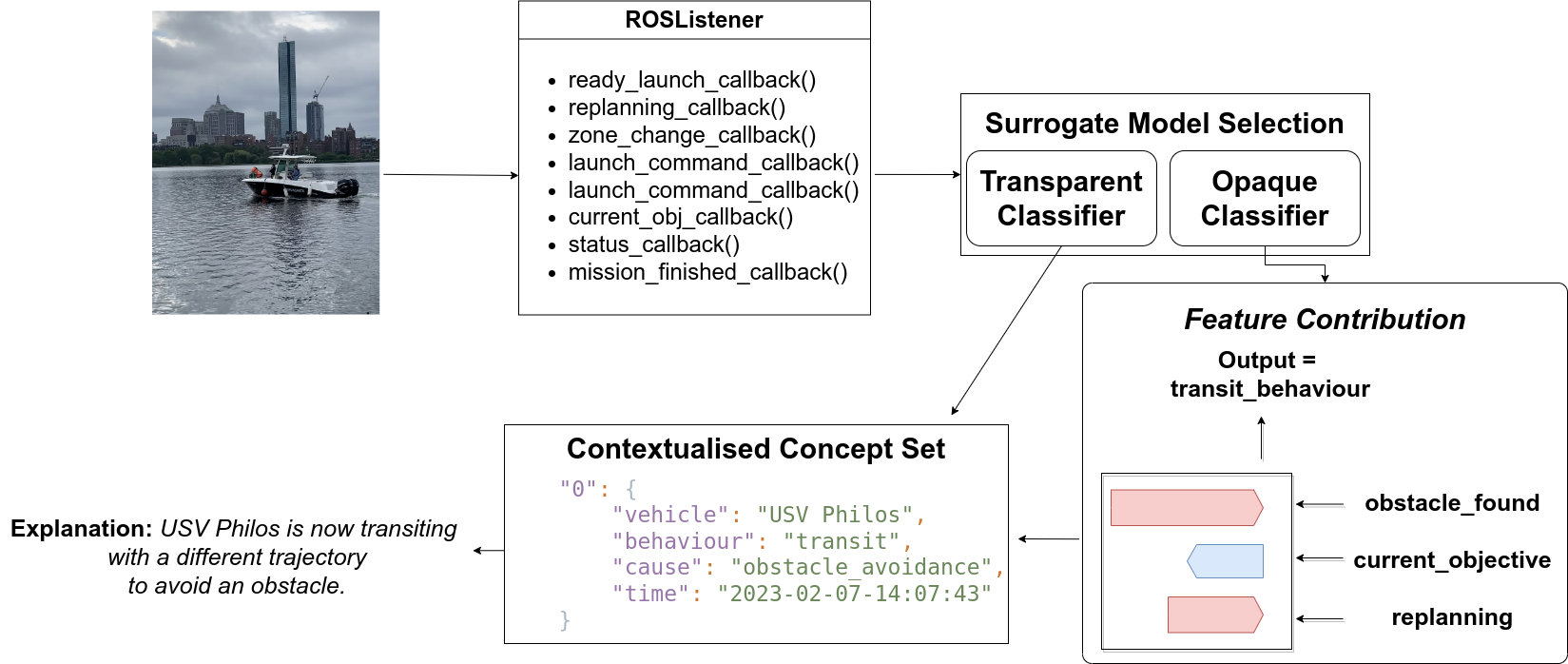}
\caption{Illustration of the proposed pipeline architecture, where a Surrogate Model approximates agent policy and Feature Contribution is estimated to detect behaviour causality. The output of this framework is an explanation with content which originates from a Contextualised Concept Set.
}
\label{fig: pipeline_architecture}
\end{figure*}

The overall pipeline architecture that goes from the autonomous vehicle to the explanation interface is illustrated in Figure \ref{fig: pipeline_architecture}. Its aim is to act as a wrapper application that does not disrupt the existing autonomy but clearly conveys the approximated policy to users. For the ROSListening component,  we found the relevant ROS topics that provide the vehicle states needed to predict exhibited behaviours (per the behaviour definition in Figure \ref{fig: empirical_decision_tree}). We then created a listener with two uses in mind: (i) data collection in simulation and (ii) online behaviour prediction during plan execution. Using the acquired data, we trained a number of classifier models as surrogate autonomy models.  These models predict which of the 6 behaviours the vehicle is exhibiting. We investigated a number of classifiers with varying transparency and compared the accuracy of the models. 

If the highest accuracy model is transparent, we would directly extract the feature contribution for each behaviour prediction. Otherwise, a post-hoc explanation method would be applied to the opaque model to derive feature contribution. Here, we examine both of these options.  Finally, a knowledge representation that contains this information is generated and fed into a rule-based natural language explanation generator that conveys the same content in a user-friendly format.  These components are described in detail below and represented in Algorithm \ref{alg: framework_steps}.  

\begin{algorithm}[!t]

\DontPrintSemicolon
\KwInput{$\mathcal{UP}$: \small{User Preference.}}
\KwInput{$\mathcal{SD}$: \small{State Dataset.}}
\KwOutput{ $\mathcal{IKR}$: \small{Intermediate Knowledge Representation.}}
\KwOutput{ $\mathcal{SV}$: \small{State Verbalisation.}}
\Begin{
 $\mathcal{UC} \leftarrow$ SelectUseCase($\mathcal{UP}$)\\
 $D \leftarrow $ SplitAndEncode($\mathcal{SD}, \mathcal{UP}$)\\
 $\mathcal{M} \leftarrow$ TrainModel($\mathcal{D}, \mathcal{UP}$)\\
 $\mathcal{E} \leftarrow$ ConfigureExplainer($\mathcal{M}, \mathcal{UP}$)\\
 $s \leftarrow $ StartExecution($\mathcal{UC}$)\\
 
\While {\textbf{not } $s$ $\leftarrow$
$\emptyset$}
  {
   $p \leftarrow M$.Predict($s$)\\
   $e \leftarrow E$.ExplainPrediction($p$)\\
   $r \leftarrow$.GenerateRepresentation($p, e$)\\
   $v \leftarrow$.VerbaliseRepresentation($r$)\\

   \If{$\mathcal{UC} \leftarrow$ isFinished}
   {
    $i$ $\leftarrow \emptyset$
   }
   \Else
   {
    $s \leftarrow $ GetNextState()\\
   }
  }
}
\caption{Explanation Framework Workflow}
\label{alg: framework_steps}
\end{algorithm}


\subsection{The Data}
\label{data_retrieval}
We collected a Behaviour Causality Dataset from 10 simulations of missions. For each simulation, we monitored eight ROS topics with a listener module corresponding to five vehicle states where S = \{\textit{ready\_plan, current\_objective, progress\_type, same\_objective, obstacle\_found}\} along with the corresponding behaviour. Each mission lasted 22.5 minutes on average and resulted in a dataset of 5056 data instances with 5 categorical features and a target value.

\subsection{Surrogate Model Training and Selection}
\label{model_selection}
After data collection, we made a comparison of various models to decide on the most suitable option for Behaviour Classification. Specifically, we tested three algorithmically explainable models, which are robust for classification with categorical features (K-Nearest Neighbours (KNN) \cite{b37}, Categorical Naive Bayes (CategoricalNB) \cite{b36} and Decision Tree\cite{b35}) and we also included two more complex models (Support Vector Machine (SVM)\cite{b38}, Multilayer Preceptron (MLP) \cite{b39}) to check if there is a significant performance difference. 

A total of 5 categorical features were given as input to each model (ready\_plan, current\_obj, progress\_type, same\_obj, obstacle\_found) to predict the current behaviour of the vessel (wait, transit, survey, hold\_position, replanned\_transit, obstacle\_avoidance). Nested cross-validation was used to select the best combination of hyperparameters for each model and to retrieve unbiased metrics indicative of each model's performance \cite{b40}. 

From the results in Table \ref{model_evaluation_table}, for the transparent models, it is clear that the Decision Tree and Categorical Naive Bayes algorithms outperform  KNN. With regards the more opaque algorithms, SVM achieved similar performance to MLP but with the latter performing slightly better at correctly classifying behaviours. As for model training and evaluation, the time needed for transparent models to do both was much shorter than for Neural Networks. Thus going forward, we decided to use the decision tree (\textit{max\_depth = 8, max\_leaf\_nodes = 15}) given that it has similar accuracy to more complex models. Furthermore, its high transparency means that it can be used to verify the validity of the surrogate framework for more opaque models, in case these are used in future use cases and datasets. 

\begin{figure}[b!]
\centering
\includegraphics[width=1.0\columnwidth]{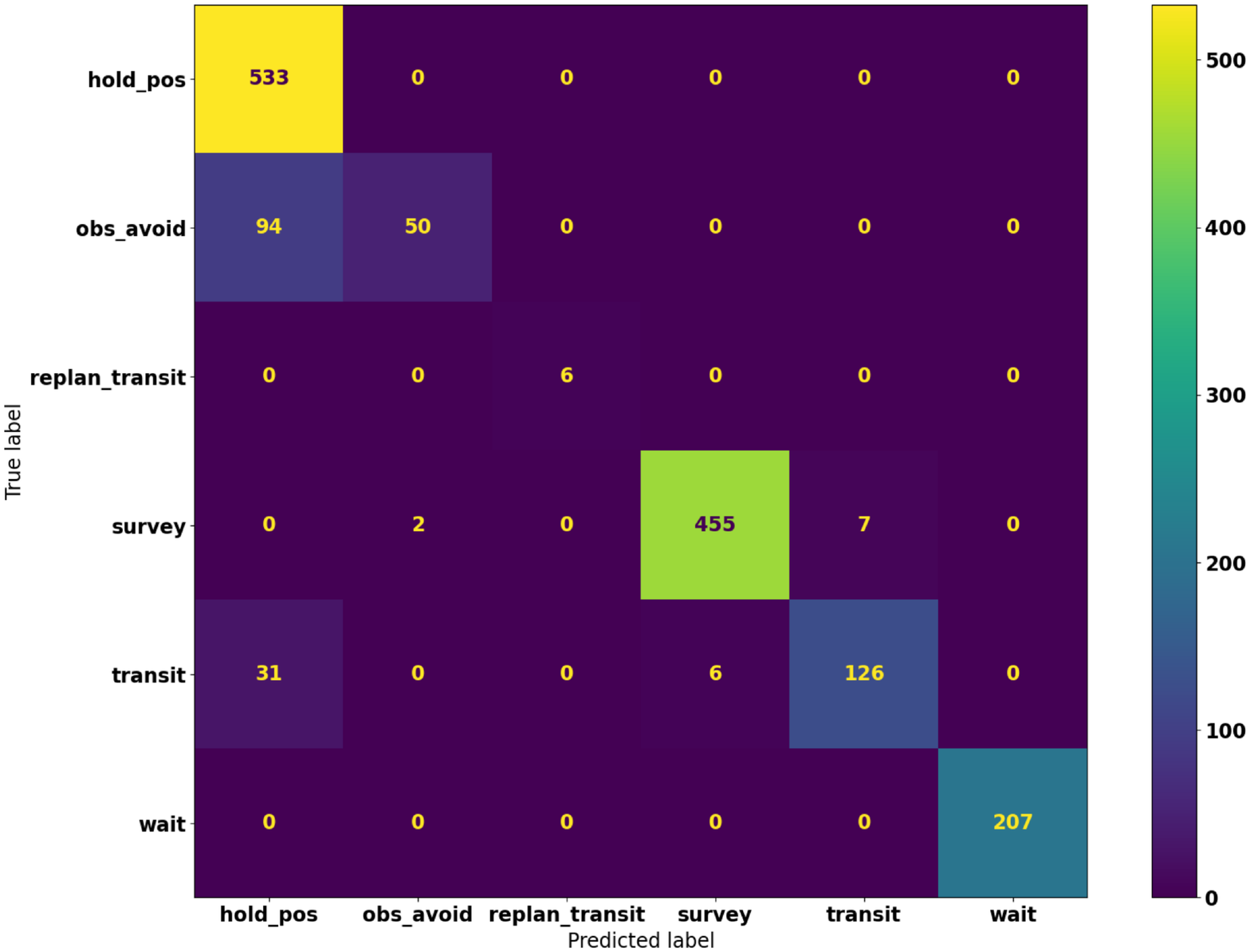}
\caption{Confusion matrix indicating classification performance per behaviour with a Decision Tree during simulations.}
\label{fig: confusion_matrix}
\end{figure}

Figure \ref{fig: confusion_matrix} provides a confusion matrix of the predictions of the decision tree per behaviour. 
For \textbf{transit}, \textbf{hold\_pos} and \textbf{survey} behaviours, there are some false classifications due to some inconsistency between the progress\_type feature and the corresponding behaviour, which indicates that an internal autonomy state could be missing. As for false classifications between \textbf{hold\_pos}, \textbf{survey} and \textbf{obs\_avoid} behaviours, we noticed that even though replanning is triggered and an obstacle is found, the vehicle finds a way to perform its objective, however, the explanation framework misses this fact probably because an internal autonomy state is missing once again.    


\begin{figure*}
     \centering
     \begin{subfigure}[b]{1.0\columnwidth}
         \centering
         \includegraphics[width=\textwidth , trim = 2cm 4cm 2cm 2cm, clip]{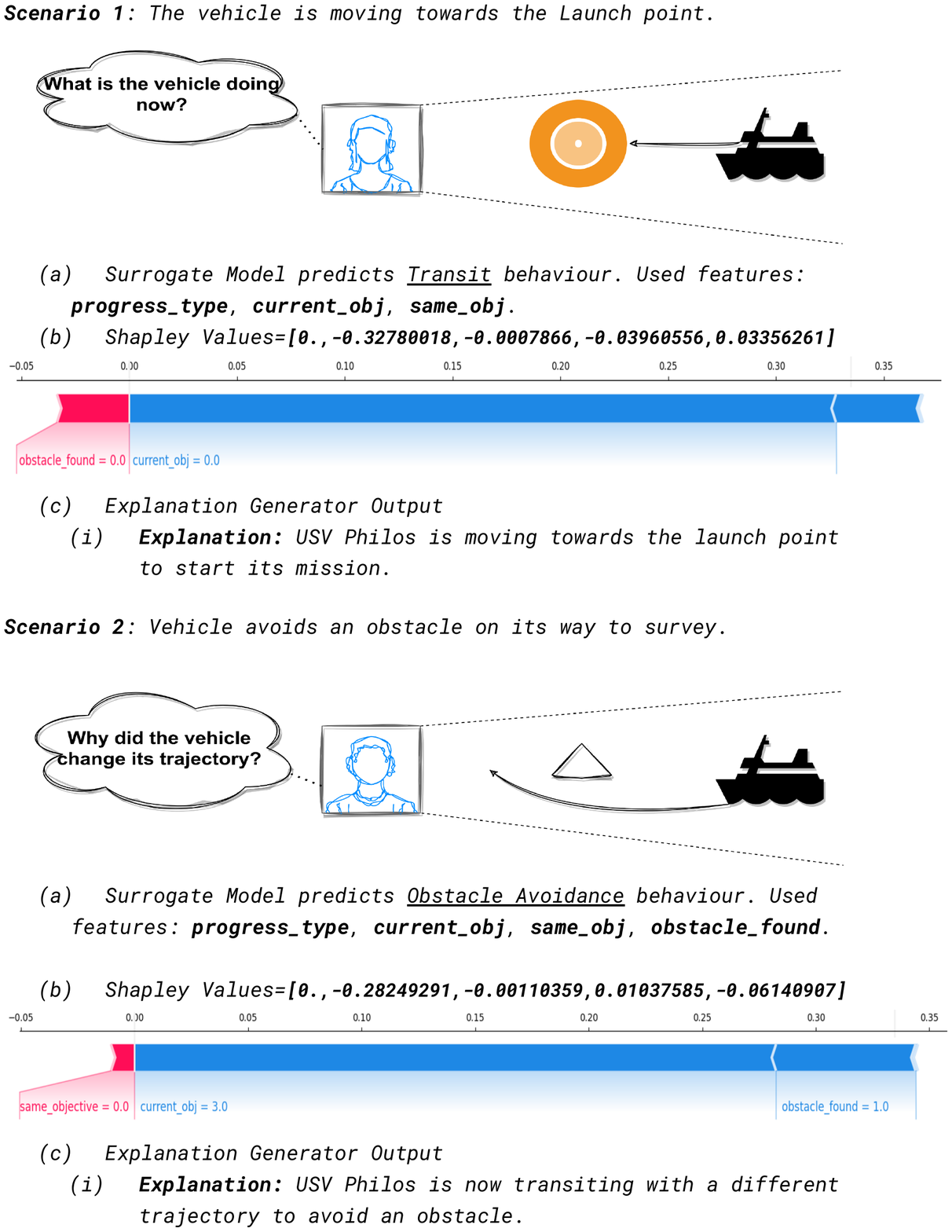}
         \label{fig:scenario1-2}
     \end{subfigure}
     \hfill
     \begin{subfigure}[b]{1.0\columnwidth}
         \centering
         \includegraphics[width=\textwidth, trim = 2cm 4cm 2cm 2cm, clip]{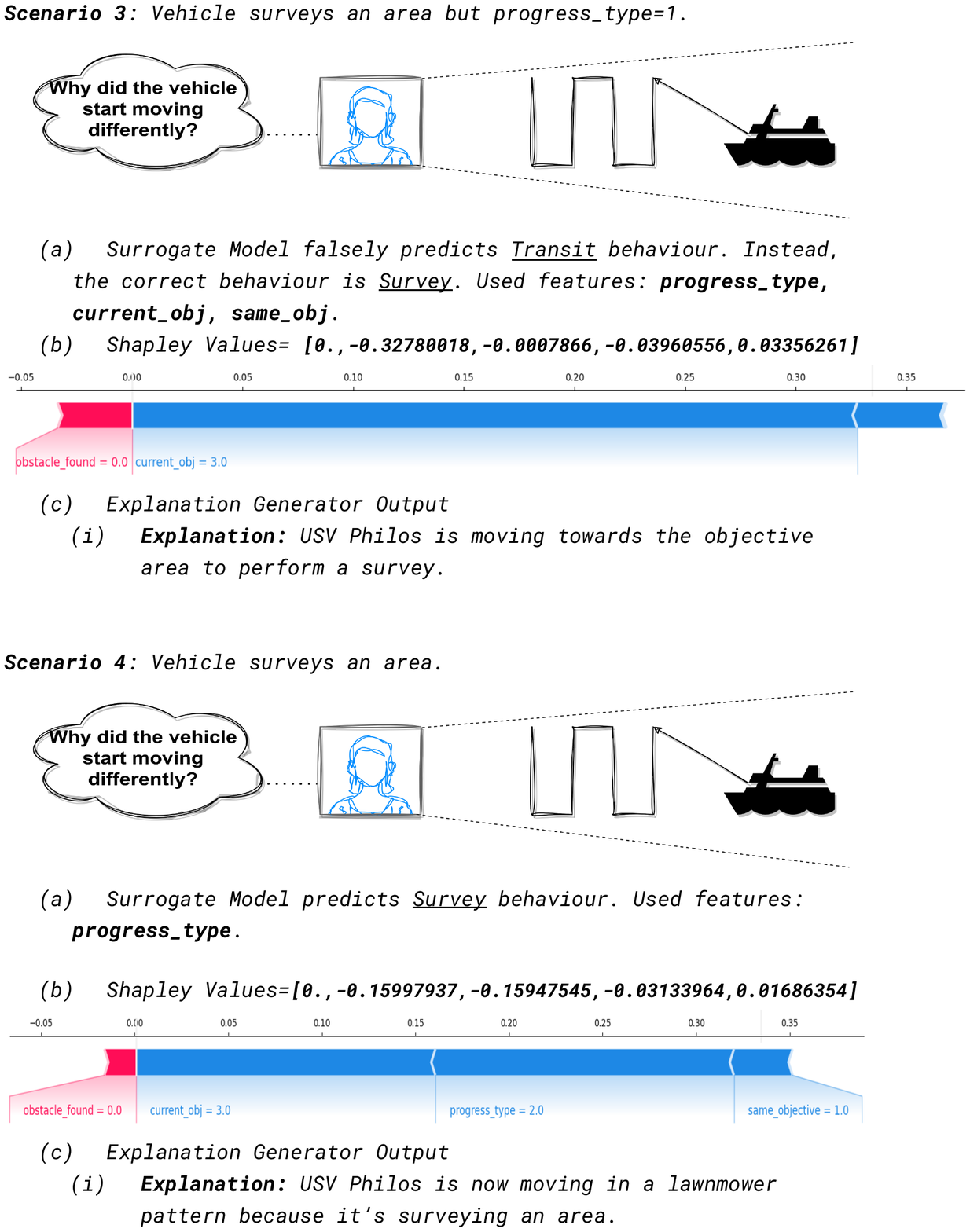}
         \label{scenario3-4}
     \end{subfigure}
        \caption{Four continuous events from a single mission along with their behaviour predictions and the corresponding explanations. Scenarios 1, 2 and 4 contain correctly predicted behaviours, while Scenario 3 demonstrates a false prediction that has been encountered.}
        \label{fig:scenarios}
\end{figure*}

\subsection{Explanation Layer}
\label{explanation_layer}

Once a trained surrogate model is in place to predict the corresponding behaviour of a vehicle state, the feature contribution for the classification of the behaviour is used as a basis for the causal reasoning explanation. 
One way to do feature contribution is to examine the trained surrogate model itself \cite{b14}. This is feasible for transparent models, such as the decision tree chosen here, but not so for more complex models such as Neural Networks. Opaque models such as Neural Networks may be needed in future applications as the complexity of the autonomy increases and the datasets grow in size. For these more complex models, an alternative is to use \textit{Shapley Values} \cite{b15}, which has been shown to be a reliable and descriptive approach. Here, we follow this latter method as a proof of concept. Each model has initially a prior belief about what the expected value will be and a Shapley value describes how a specific feature creates the difference between the expected and actual values (\begin{math} E(x) - f(x)\end{math}).


\subsection{Knowledge Representation and Explanation Generation}
\label{knowledge_representation}

The final two components of the pipeline use, as input, the prediction of the surrogate model along with the Shapley values estimated by the feature contribution estimator. Based on the importance of each feature towards a prediction, behaviour causality is inferred and this knowledge is represented with contextualised concept sets. Contextualisation is incorporated with the use of key-value pairs, as opposed to simple triplets to indicate the role of each value. The end result is  a knowledge base with \begin{math}(vessel, behaviour, causality, time)\end{math} sets, which describe the sequence of behaviours exhibited by the robot in JSON format. An example of a contextualised concept set can be found in Figure \ref{fig: pipeline_architecture}, where the current behaviour (Transit) and its trigger (Obstacle) can be distinguished. This entry indicates that the current transit behaviour has a modified trajectory that goes around the obstacle to avoid collision.  With regards to natural language generation, for each new entry in the Knowledge Base, the key-value pairs are passed to a \textit{Surface Realiser}, which produces an explanation that has been syntactically checked with SimpleNLG \cite{b34}.


\begin{table*}[ht]
\centering
\begin{tabular}{|c|c|c|c|c|c|c|}
	\hline
	\textbf{Models} & \textbf{Accuracy} & \textbf{Precision} & \textbf{Recall} & \textbf{F1-Score} & \textbf{Fit Time} & \textbf{Score Time}\\
	\hline
    Decision Tree & \textbf{0.8981} & \textbf{0.9464} & 0.8498 & \textbf{0.8712} & 25.0936 & 0.0025\\
    \hline
    CategoricalNB & 0.8247 & 0.8701 & \textbf{0.8616} & 0.8379 & \textbf{0.1127} & \textbf{0.0019} \\
    \hline
    KNN & 0.6655 & 0.7806 & 0.8291 & 0.6953 & 4.8554 & 0.0721 \\
    \hline \hline
    SVM & 0.8846 & 0.9163 & 0.8378 & 0.8535 & 14.9298 & 0.0547 \\
    \hline
    Multilayer Perceptron (MLP) & \textbf{0.8987} & 0.9459 & 0.8496 & 0.8707 & 147.8816 & 0.0075 \\
    \hline
\end{tabular}
\caption{Classification performance metrics across five different models derived with nested cross-validation on simulation data. Transparent models are at the top part of the table and opaque ones are at the bottom. Training and score times (in seconds) are also included to indicate the effort for acquiring a surrogate model in other similar use cases.}
\label{model_evaluation_table}
\end{table*}

\section{Results and Discussion} \label{results}
With regards \textbf{RQ1} and \textbf{RQ2}, as discussed above and in Table \ref{model_evaluation_table}, the surrogate models have accuracies for behaviour prediction of around 90\%. This could further be further improved by training on more data both in simulation and with real vehicles.

Even with this accuracy, we observed accurate explanations as give illustrated in Figure \ref{fig:scenarios}.  In this figure, we present four continuous scenarios from a single mission along with the explanations which were generated.  \textit{Scenario 1} involves a vessel moving to the launch point to retrieve the relative positions of the objective areas and begins working on each objective. In this case, the surrogate model correctly predicts the current behaviour by using the \textbf{progress\_type}, \textbf{current\_obj} and \textbf{same\_obj} features. To validate the results, we also calculated the Shapley values, which highlight the \textbf{current\_obj} as the main contributor. In \textit{Scenario 2}, while the vehicle is moving from the launch point to the survey area, it encounters an obstacle and avoids it by changing its trajectory. Behaviour prediction was also successful in this case, with the model utilising \textbf{progress\_type}, \textbf{current\_obj}, \textbf{same\_obj} and \textbf{obstacle\_found} features. In \textit{Scenario 3}, a false explanation is generated, due to the value of \textbf{progress\_type} even though the survey has already started. Here, the features used by the surrogate model are, \textbf{progress\_type}, \textbf{current\_obj} and \textbf{same\_obj}, while SHAP only attributes this prediction to the current objective. As for \textit{Scenario 4}, where the vessel performs a survey, the surrogate model can immediately detect the new behaviour thanks to its unique \textbf{progress\_type}, while SHAP adds to the causality both \textbf{current\_obj} and \textbf{same\_obj}, which seem reasonable causes in this case. An informal evaluation has been conducted but a formal subject evaluation is future work. 

\subsection{Going from Simulation to Real Trial}
\begin{figure}[!t]
\centering
\includegraphics[width=1.0\columnwidth]{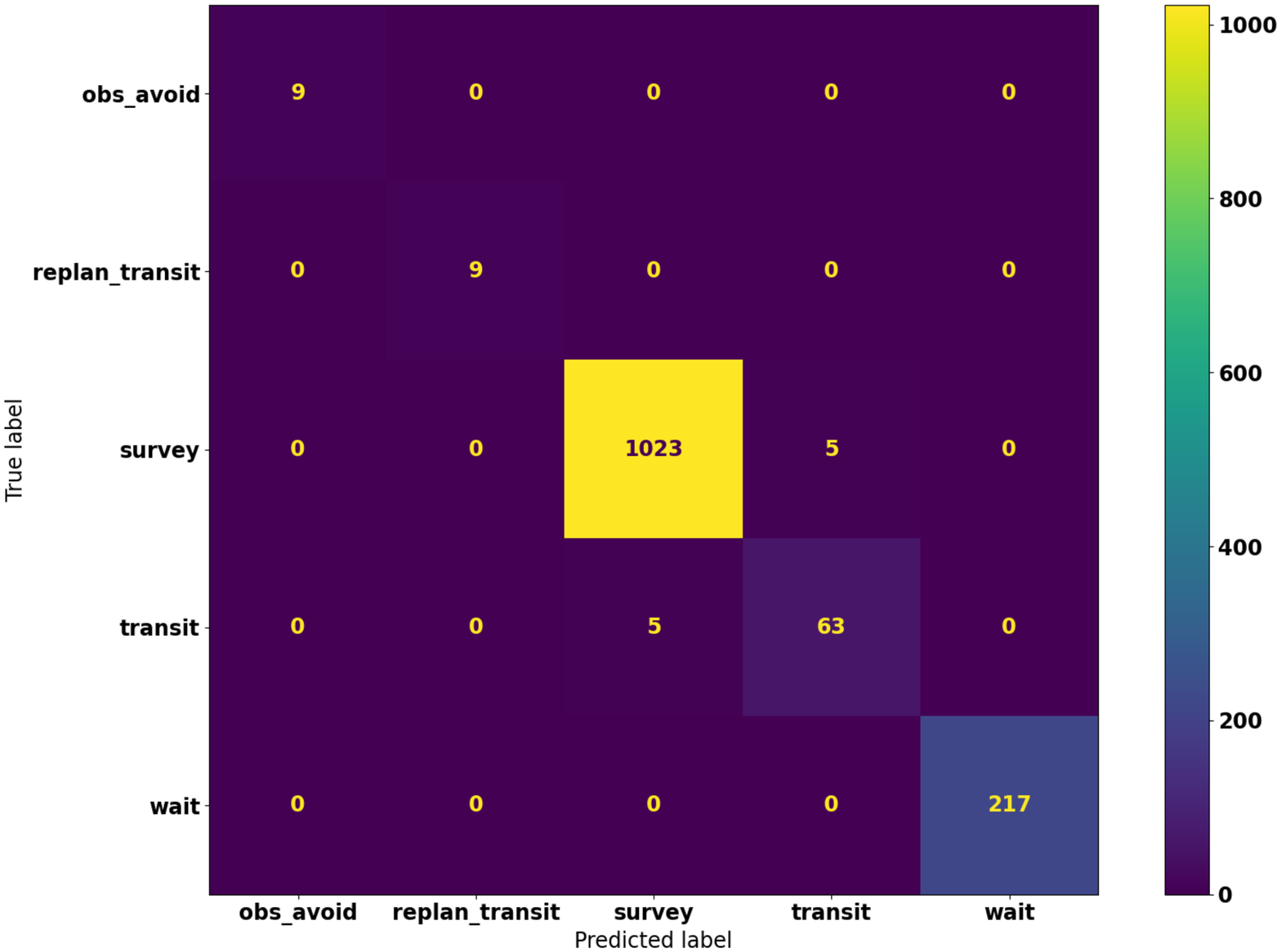}
\caption{Confusion matrix indicating classification performance per behaviour with a Decision Tree during the real trial.}
\label{fig: trial_confusion_matrix}
\end{figure}

For \textbf{RQ3}, a real trial took place with two Unmanned Surface Vehicles collaborating to complete a survey, while obstacles appeared in the dynamic environment. As a result, we tested the surrogate model on a separate trial test set consisting of 1331 instances with 5 features and corresponding behaviours. Looking at the overall performance, an accuracy of 99\% was achieved showing the capability of the surrogate model to comprehend the autonomous behaviour activations. See Figure \ref{fig: trial_confusion_matrix} for the confusion matrix for this test. The only errors observed are false classifications between the \textbf{survey} and \textbf{transit} behaviours, which we attribute to a potential missing vehicle state since the \textbf{progress\_type} could not fully indicate the transition between these two behaviours. Finally, the \textbf{hold\_position} behaviour is missing from Figure \ref{fig: trial_confusion_matrix}, because this objective was not used during the trial for practical reasons. 


\section{Conclusion and Future Work}
With this work, a framework for approximating behaviour activations and replanning of an autonomous agent with classification models has been introduced. Our approach is capable of discovering the causality of autonomous decisions with the estimation of feature contribution for each action prediction. The main advantage of this framework is the storage of information in generic knowledge representations such as concept sets which can be later leveraged to produce user-friendly modalities such as natural language explanations. Another advantage is that the framework is agnostic to the autonomy model, making it reusable in different domains. Moving forward, we plan on extending the functionality of this framework and investigating data-driven language explanations such as large language models in order to stochastically map knowledge representations to informative natural language explanations about robotic behaviour.
Further evaluation of explanations is also required to examine the capacity of our approach to disambiguate robotic behaviours.

\section*{Acknowledgment}
We would like to thank MIT's AUV Lab and Laurence Boe from SeeByte Ltd for their assistance with the simulator and the real trial. This work was also funded and supported by the EPSRC Prosperity Partnership (EP/V05676X/1), the UKRI Node on Trust (EP/V026682/1), EPSRC CDT on Robotics and Autonomous Systems (EP/S023208/1), and Scottish Research Partnership in Engineering.


\vspace{12pt}
\end{document}